\documentclass[sn-mathphys-num]{sn-jnl}

\usepackage{graphicx}%
\usepackage{multirow}%
\usepackage{amsmath,amssymb,amsfonts}%
\usepackage{amsthm}%
\usepackage{mathrsfs}%
\usepackage[title]{appendix}%
\usepackage{xcolor}%
\usepackage{textcomp}%
\usepackage{manyfoot}%
\usepackage{booktabs}%
\usepackage{algorithm}%
\usepackage{algorithmicx}%
\usepackage{algpseudocode}%
\usepackage{listings}%
\newcommand{\framedtext}[1]{%
\par%
\noindent\fbox{%
    \parbox{\dimexpr\linewidth-2\fboxsep-2\fboxrule}{#1}%
}%
}

\begin{document}

\title[Article Title]{Large Language Models for Ingredient Substitution in Food Recipes using Supervised Fine-tuning and Direct Preference Optimization}


\author[1]{\fnm{Thevin} \sur{Senath}}\email{thevin.19@cse.mrt.ac.lk}

\author[1]{\fnm{Kumuthu} \sur{Athukorala}}\email{kumuthu.19@cse.mrt.ac.lk}

\author[1]{\fnm{Ransika} \sur{Costa}}\email{rumal.19@cse.mrt.ac.lk}
\author*[2]{\fnm{Surangika} \sur{Ranathunga}}\email{s.ranathunga@massey.ac.nz}

\author[3,4]{\fnm{Rishemjit} \sur{Kaur}}\email{rishemjit.kaur@csio.res.in}

\affil[1]{\orgdiv{Department of Computer Science and Engineering}, \orgname{University of Moratuwa}, \orgaddress{\street{Katubedda},  \postcode{10400},  \country{Sri Lanka}}}

\affil*[2]{\orgdiv{School of Mathematical and Computational Sciences}, \orgname{Massey University}, \orgaddress{\city{Auckland}, \postcode{0632},  \country{New Zealand}}}

\affil[3]{\orgname{CSIR-Central Scientific Instruments Organisation}, \orgaddress{\street{Sector-30C}, \city{Chandigarh}, \postcode{160030}, \country{India}}}
\affil[4]{\orgname{Academy of Scientific and Innovative Research (AcSIR)}, \orgaddress{\city{Ghaziabad}, \postcode{201002}, \state{U.P.}, \country{India}}}


\abstract{In this paper, we address the challenge of recipe personalization through ingredient substitution. We make use of Large Language Models (LLMs) to build an ingredient substitution system designed to predict plausible substitute ingredients within a given recipe context. Given that the use of LLMs for this task has been barely done, we carry out an extensive set of experiments to determine the best LLM, prompt, and the fine-tuning setups. We further experiment with methods such as multi-task learning, two-stage fine-tuning, and Direct Preference Optimization (DPO). The experiments are conducted using the publicly available Recipe1MSub corpus. The best results are produced by the Mistral7-Base LLM after fine-tuning and DPO. This result outperforms the strong baseline available for the same corpus with a Hit@1 score of 22.04. Thus we believe that this research represents a significant step towards enabling personalized and creative culinary experiences by utilizing LLM-based ingredient substitution.}

\keywords{Natural Language Processing, Parameter-Efficient Fine-Tuning, Multi-Task Learning, Two-Stage Fine-Tuning, Direct Preference Optimization,  recipe personaization, Mistral}



\maketitle

\section{Introduction}\label{sec1}

Food recipes from diverse cultures and regions are spreading rapidly on the Internet, igniting curiosity to explore new culinary experiences. This exposure to various cuisines broadens our culinary horizons and encourages experimentation with different dishes at home. However, preparing new recipes often involves personalizing them to accommodate ingredient availability, cost, dietary restrictions, lifestyle choices, allergies, or personal taste preferences. This customization typically involves substituting certain ingredients with alternatives. Yet, finding suitable substitutes that maintain the intended flavor profile can be challenging, as the effectiveness of substitutions varies depending on the recipe and context. For instance, while oil may work well as a substitute for butter in frying chicken, it might not yield the same results in cake baking. Personalizing recipes through ingredient substitution has the potential to enhance culinary experiences by accommodating factors such as cost, availability, and dietary needs. 

Past research has experimented with various Natural Language Processing (NLP) techniques for this task, and several pre-trained Language Models (PLMs) based on Word2Vec~\cite{mikolov2013efficient}  and ~BERT\cite{devlin-etal-2019-bert} have been proposed~\cite{pellegrini2021exploiting, lawo2020supporting}. \citet{fatemi2023learning}'s pioneering work \emph{Graph-based Ingredient Substitution Module (GISMo)} uses recipe-specific context and generic ingredient relational information encoded as a graph simultaneously to predict suitable substitutions.

Large Language Models (LLMs), combined with advanced prompting techniques, have emerged as the current de-facto solution for NLP tasks in various domains. However, their application in ingredient substitution remains under-explored, and we are aware of only one very recent work~\cite{rita2024optimizing}. Therefore, this paper aims to implement an LLM-based ingredient substitution solution that considers the recipe's context for accurate ingredient replacement. Given that LLMs have only used once for the task of ingredient substitution before, we first carry out a set of experiments to determine the best publicly available LLM and the best prompt, using zero-shot { and few-shot testing.}  Experiments are conducted using the publicly available Recipe1MSub~\cite{fatemi2023learning} benchmark dataset. Mistral7B-base model emerged as the best candidate for the task. Subsequently, we perform supervised fine-tuning (SFT) of this selected LLM. We experiment with multiple Parameter Efficient Fine-tuning (PEFT) techniques and identify  {QLoRA} to be the best PEFT technique. We carry out further experiments using two-stage fine-tuning, multi-task fine-tuning, as well as Direct Preference Optimization (DPO) techniques. Our best model that employs both SFT and DPO outperforms the strong baseline of GISMO, with a Hit@1 score of 22.04.

To summarize, the contributions of this paper are as follows:

\begin{itemize}
    \item Investigation into the best LLM prompting patterns for ingredient substitution and introduction of a new prompt that effectively makes use of the information in the recipe.
    \item Extensive experiments to determine the best LLM as well as optimization strategies for the ingredient substitution task.
    \item Presents a system that incorporate SFT and DPO, for the ingredient substitution task
\end{itemize}

The rest of the paper is organized as follows. Section 2 discusses past work related to ingredient substitution and LLMs. Section 3 discusses our methodology. This includes a detailed discussion on the prompt development, LLM fine-tuning as well as performance optimization strategies. Section 4 presents the dataset and the experiment setup,  and section 5 presents the results. Finally, section 6 concludes the paper with limitations of the research and a look into future research directions.

\section{Related work}
\label{sec:related}
\subsection{Ingredient Substitution}
Most of the existing approaches for ingredient substitution are based on textual recipe information and statistics~\cite{boscarino2014automatic, achananuparp2016extracting}. In recent years, researchers have started to explore NLP techniques to address this ingredient substitution task using PLMs such as BERT, R-BERT \cite{wu2019enriching} and Word2Vec.~ \citet{pellegrini2021exploiting} introduced two PLMs, namely food2vec and foodBERT, to obtain ingredient embeddings from recipe instructions and user comments. These learned embeddings were applied to the context of ingredient substitution in dietary use cases. \citet{lawo2020supporting} introduced ingredient2Vec, a Word2Vec-based model, to develop a neural network model for suggesting vegan substitutes for omnivorous ingredients.

A recent approach to address this task is the \emph{Graph-based Ingredient Substitution Module (GISMo)} \cite{fatemi2023learning}, which uses recipe-specific context and generic ingredient relational information encoded as a graph simultaneously to predict suitable substitutions. {GISMo uses recent advancements in graph neural networks (GNNs) to capture interactions between ingredients within learned ingredient embeddings. These embeddings are then refined with contextual information from a given recipe, enabling the prediction of plausible ingredient substitutions.} Additionally, the authors have introduced a benchmark known as Recipe1MSubs, featuring substitution pairs linked to the Recipe1M dataset \cite{salvador2017learning}, with standardized splits, evaluation metrics, and baselines.

Some research has used LLMs and Vision-Language pre-trained models for recipe generation and retrieval~\cite{menon2024generating,wahed2024fine,mohbat2024llava,kamatchi2024comparative}. Parallel to our work,~\citet{rita2024optimizing} experimented with LLMs for the ingredient substitution task. However, they have only employed SFT, and there has been no study on identifying the most effective prompt pattern or the recipe context. 

\subsection{Large Language Models}
\subsubsection*{Introduction}
Pre-trained Language Models (PLMs) derived by training the Transformer neural architecture~\cite{vaswani2017attention} with large corpora have revolutionized the field of NLP. These PLMs can be either encoder-based, decoder-based or encoder-decoder based. Out of these, decoder-based PLMs have gained prominence with the success of commercial products such as ChatGPT. Recent decoder-based models have been trained using much larger corpora compared to the early-day PLMs, thus earning them the term LLMs. Llama~\cite{touvron_Llama_2023}, Mistral~\cite{jiang2023mistral} and Gemma~\cite{team2024gemma} are some of the commonly used publicly available LLMs that have shown comparable performance to their commercial counterparts.

These LLMs are trained using large amounts of raw text crawled from the web, using the \textit{next word prediction} self-supervised learning objective. This training process is termed \textit{pre-training}, and the resulting models are referred to as \textit{base} models. In general, we can represent a base model consisting of $M$ parameters where $\theta = [\theta_0, \theta_1, ..., \theta_M]$ .  In order to utilize such base models for down-stream NLP tasks, they have to be further fine-tuned (supervised fine-tuning (SFT)) using task-specific data. The fine-tuning process can be denoted as a transition of these $M$ parameters to $\hat{\theta} = [\hat{\theta_0}, \hat{\theta_1}, ..., \hat{\theta_M}]$. This is done using training data in the form of instruction-answer pairs (together, these constitute the \textit{prompt}), thus this SFT stage is referred to as \textit{instruct tuning}~\cite{longpre2023flan}. Most LLMs today come in both base and instruct forms. A base LLM or an instruct-tuned version can be used for new NLP tasks, without any task-related SFT. This is termed \textit{zero-shot prompting}. A better approach is \textit{in-context learning}~\cite{dong2022survey}, where a few examples of the same task are included in the prompt, for the LLM to learn the task by referring to them.  This is also termed \textit{few-shot learning }or \textit{few-shot prompting}.

\subsubsection*{Prompt Engineering}
\label{prompt_lit}
The success of LLMs largely depends on the prompt provided to the LLM.  Researchers have introduced various prompting patterns and techniques that significantly improve the performance of LLMs. \citet{white2023prompt}, mention several prompt patterns that can be used to get more accurate results using an LLM. Using prompt patterns, the context of the original problem can be provided to the LLM. The Output Automata Pattern, Template Pattern, Persona Pattern, and Context Manager Pattern are some of the prompt patterns that are commonly used~\cite{white2023prompt}.

\subsubsection*{Parameter Efficient Fine Tuning (PEFT)}
Fine-tuning LLMs is computationally and memory intensive, leading to the development of PEFT techniques that enhance memory and computational efficiency.~\citet{hu2021lora} introduced Low-Rank Adaptation (LoRA), where the layers of the LLM remain fixed, and a trainable low-rank decomposition matrix (composed of two linear projections) is added to each layer. In other words, in LoRA, the transition of parameters can be encoded by a much smaller sized set of parameters $\Phi = [\Phi_0, \Phi_1, ..., \Phi_N]$ where $N << M$. By decomposing the update into smaller matrices, LoRA significantly reduces the number of parameters needing training.  Quantized LoRA (QLoRA)~\cite{dettmers2024qlora} uses quantization strategies on LLM weights and trains LoRA modules on this quantized representation. GaLore~\cite{zhao2024galore} is aimed at reducing VRAM requirements by optimizing the training process of model parameters rather than reducing the number of parameters. It employs two main strategies, Gradient Low-Rank Projection and Per-Layer Weight Updates. {Prompt Tuning}~\cite{lester2021power}, {BitFit}~\cite{zaken2021bitfit}, {IA3}~\cite{liu2022few}, {AdaLoRA}~\cite{zhang2023adalora} and {ProPELT}~\cite{zeng2023one} are some of the other PEFT techniques that are used to fine-tune LLMs.

\subsubsection*{Direct Preference Optimization}

Preference Optimization is a technique employed to further boost the performance of LLMs after they undergo SFT. Reinforcement Learning from Human Feedback (RLHF)~\cite{bai2022training} is an effective online preference optimization technique. However, the use of RLHF is cumbersome, as two models have to be trained in parallel. {Direct Preference Optimization (DPO)~\cite{rafailov2023direct}, which is an offline policy optimization technique, is an alternative to RLHF and avoids the need for a reward model. DPO operates on preference data, which consists of triplets in the form of (prompt, chosen answer, rejected answer). This type of data is similar to that used in RLHF, where it would typically be used to train a reward model for subsequent reinforcement learning. However, in DPO, there is no reinforcement learning step; instead, the model is optimized directly on this preference data.}
At the beginning of the fine-tuning process, an exact copy is done for the LLM that is being trained, and its trainable parameters are frozen. For each data point, the chosen and rejected responses are scored by both the trained and the frozen language model. This score is the product of the probabilities associated with the desired response token for each step.

{
DPO optimizes the model parameters so that, for each input, the probability of producing the chosen output becomes higher than the probability of the rejected output. This adjustment is made by calculating a loss based on the ratio of probabilities between the preferred and rejected responses relative to a reference model. }
\section{Methodology}\label{sec3}
\begin{figure}[!h]
    \centering
    \caption{Flow diagram of the methodology. For each step, the LLMs used for the experiments are indicated.}
    \includegraphics[width=0.5\textwidth]{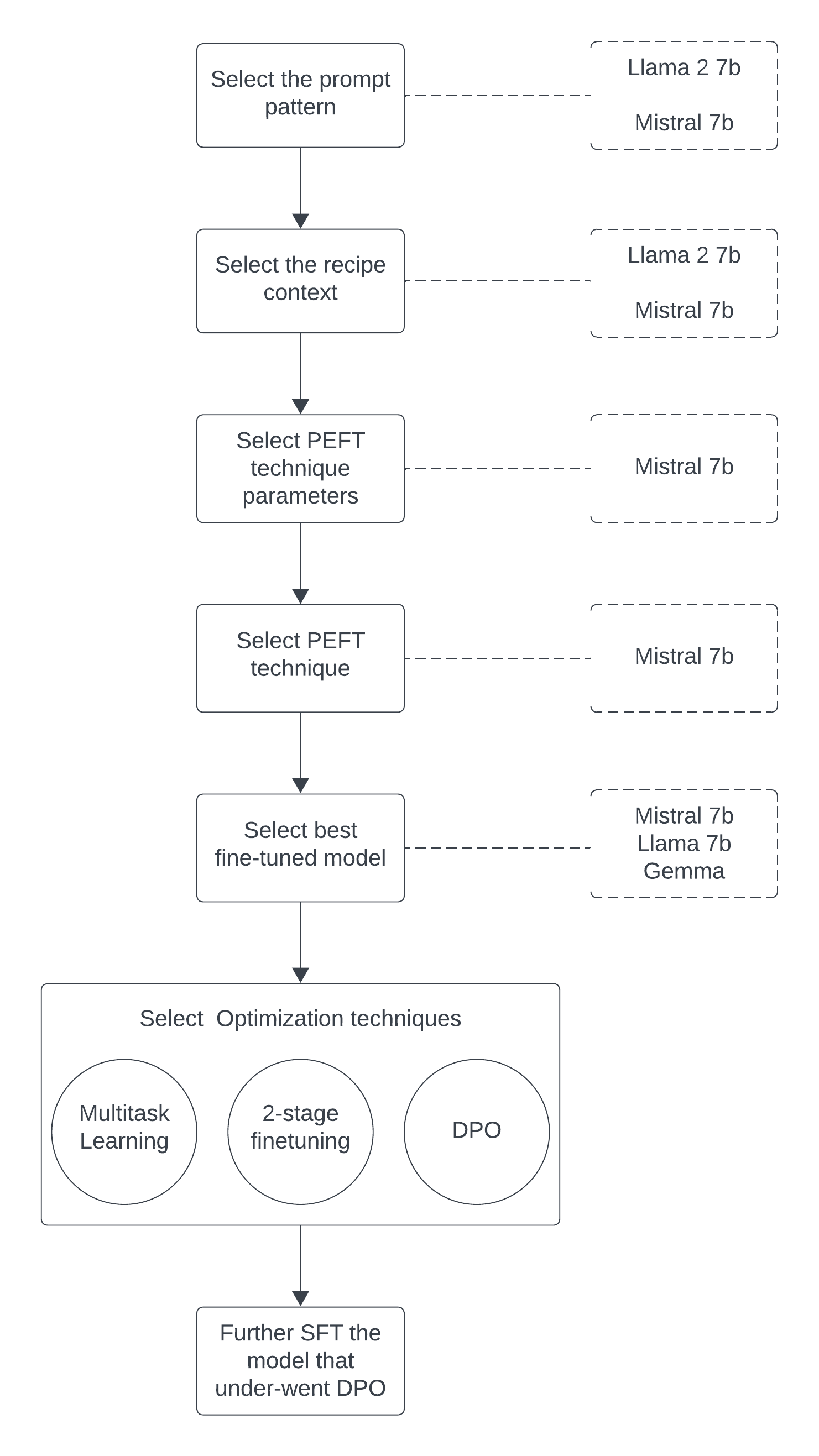}
    \label{fig:experiment}
\end{figure}

We followed a step-wise approach in designing our methodology as shown in Figure~\ref{fig:experiment}. We describe each of these steps in the subsequent sections. As further discussed in the sub-sections, we considered multiple LLMs for experiments. However, in order to keep the experiment space under a manageable level, we had to resort to a subset of LLMs for some of the experiments.

\subsection*{Prompt Selection}

As mentioned in Section~\ref{prompt_lit}, the prompt presented to the LLM plays a major role in getting the expected output from the LLM. Therefore we experiment with several state-of-the-art prompt patterns suggested by previous research. {Table \ref{table:prompt_patterns} contains the different prompt patterns that are used in the research.}
\begin{table}[h!]
\centering
\caption{Prompt patterns with examples from the ingredient substitution task}
\begin{tabular}{|p{3cm}|p{5cm}|p{5cm}|}
\hline
\textbf{Prompt Pattern} & \textbf{Description} & \textbf{Prompt} \\
\hline
\textbf{Persona Pattern} & Act as a specific persona and provide outputs that such a persona would. & As a master chef, your culinary prowess knows no bounds. \\
\hline
\textbf{Template Pattern} & To ensure an LLM’s output follows a precise template in terms of structure. & Follow the instructions below and suggest the best substitute for the given ingredient.\\
\hline
\textbf{Context Manager Pattern} & To focus the conversation on specific topics or exclude unrelated topics from consideration. & Your ability to flawlessly cook any dish is unparalleled. Even when faced with a missing ingredient, you effortlessly identify the perfect substitute \\ 
\hline
\end{tabular}
\label{table:prompt_patterns}
\end{table}

\subsection*{Prompt Selection with Different Contexts}

The main objective of our research is finding a substitute for an ingredient in a recipe. To achieve this, we need to determine the specific recipe details that need to be provided to the LLM through the prompt. The following five different contexts related to recipes are incorporated into the selected prompt pattern. 

\begin{itemize}
    \item Providing the source ingredient only
    \item Providing the source ingredient along with the recipe title
    \item Providing the source ingredient along with the recipe ingredients
    \item Providing the source ingredient along with the both recipe title and recipe ingredients
    \item {Providing the source ingredient along with the both recipe title and cooking instructions}
\end{itemize}

As further discussed in Section~\ref{sec:eval}, the optimal result was obtained with the source ingredient and the recipe title. Figure~\ref{fig:prompt} shows the prompt that derived the best results. 

\begin{figure}[htbp]
\caption{The prompt that provided the best results}
\framedtext
{
\fontsize{7.5}{8.5}\selectfont
[INST] <{}<SYS>{}>As a master chef, your culinary prowess knows no bounds. Your ability to flawlessly cook any dish is unparalleled.Even when faced with a missing ingredient, you effortlessly identify the perfect substitute. <{}</SYS>{}>\\

Follow the instructions below and suggest the best substitute for the given ingredient.\\

Instructions:\\
- Do not provide the same ingredient as above as the substitutes.\\\
- Give only one ingredient.\\
- Avoid giving explanations.\\
- Only provide the name of the ingredient.\\
- Give the output as a numbered point.\\

Dish: {recipe title} \\
Ingredient: {ingredient name}

[/INST]
}
\label{fig:prompt}
\end{figure}





\subsection*{Model Selection and Fine-Tuning}
\label{subsec:finetune}

Given that LLMs are the current de-facto solution in NLP, numerous LLMs have been introduced, and it is challenging to determine the best LLM for a given task simply by considering model properties. Therefore, we conduct extensive research on several open-source LLMs, ultimately selecting those recognized as the most recent and highest performing. Specifically, {\textbf{Llama-2 7b}, \textbf{Llama-2 7b Chat}, \textbf{Mistral 7b}, \textbf{Mistral 7b Instruct} and \textbf{Gemma 7b Instruct}} are chosen for our research. First, the selected LLMs are evaluated using zero shot and few-shot (one-shot) testing.

Using an LLM as it is under the zero-shot or few-shot setup is not optimal because it is not specifically trained on food domain data. As fine-tuning an LLM is essential for better accuracy, SFT is performed on the selected LLMs. As the first step, instruct-tuning datasets are created following the standard prompt templates of each LLM. As performing a full fine-tuning requires massive hardware, we employ PEFT techniques. Specifically, we experiment with three PEFT techniques, namely LoRA, QLoRA and GaLore.

\subsection*{Optimising the Results of the Fine-tuned Model}

Different optimization techniques are applied on the fine-tuned LLM.

\paragraph*{Two Stage Fine-Tuning}
The prompt shown in Figure~\ref{fig:prompt} makes use of only the recipe title. While this proves to be the best-performing prompt, it does not adequately expose the LLM to food recipe related knowledge. In order to provide more knowledge on food recipes, we create a  new task using the food recipe data. In other words, we convert the recipe data into question-answer pairs and use them to instruct-tune the LLM. The question is on the set of ingredient to prepare a given dish. {The prompt we use to create the dataset to fine-tune the LLM with recipe data is shown in Figure \ref{fig:recipe_data_prompt}. Afterwards, the resulting model is further fine-tuned with data for the ingredient substitution task, resulting in two-stage fine-tuning.

\begin{figure}[htbp]
\caption{The prompt used to fine-tune the LLM with question-answer pairs on the ingredients needed to prepare a dish}

\framedtext
{
\fontsize{7.5}{8.5}\selectfont
[INST] What are the ingredients we need to make Cool 'n Easy Creamy Watermelon Pie? [/INST] \\
To make Cool 'n Easy Creamy Watermelon Pie you need boiling water, cool whip, seedless watermelon, graham cracker crust.
}
\label{fig:recipe_data_prompt}
\end{figure}

\paragraph*{Multi Task Fine-tuning}

Multi-task fine-tuning involves training the model to simultaneously perform multiple tasks, leveraging shared knowledge and representations across different domains. In this approach, the model is fine-tuned with both ingredient substitution task, as well as the new task we introduced in the previous section.

\paragraph*{Direct Preference Optimization (DPO)}
To implement DPO, we construct a preference dataset from training data that initially yielded undesirable or incorrect responses to ingredient substitution prompts. Each sample is structured as a triplet in the form of (prompt, chosen answer, rejected answer) as shown in Table \ref{table:DPO Preference Dataset}. The model is then directly optimized on this preference data by adjusting model weights to increase the probability of generating responses aligned with the chosen answer.} Following DPO, we further fine-tune the optimized model again for the ingredient substitution task. 

\begin{figure}[!h]
    \centering
    \caption{System Architecture}
    \includegraphics[width=1\textwidth]{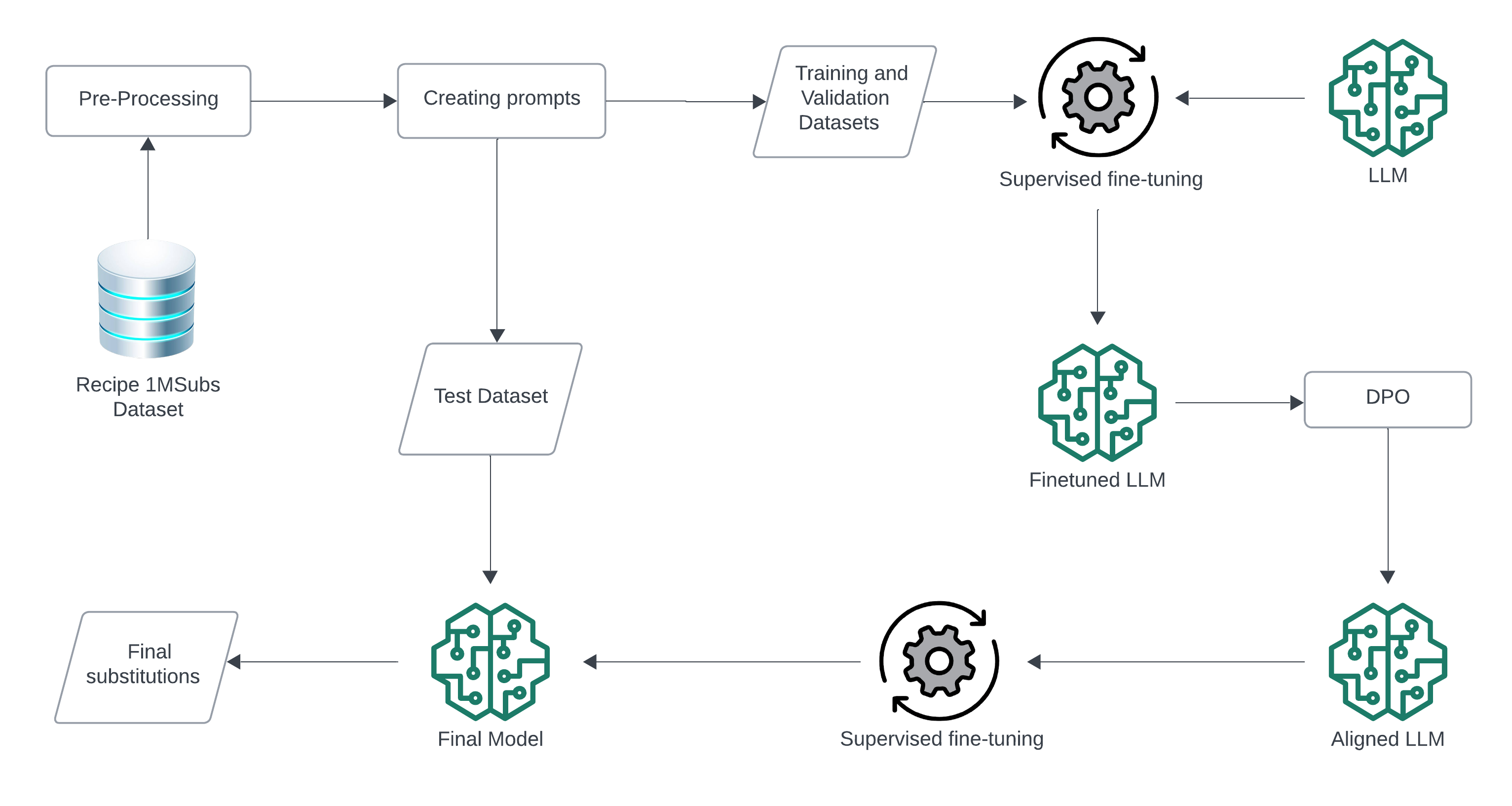}
    \label{fig:methodology}
\end{figure}

As further discussed in Section~\ref{sec:eval}, two-stage fine-tuning and multi-task fine-tuning did not result in any gains. Consequently, the model that employs SFT and DPO emerged as the best solution. The architecture of this final system is shown in Figure~\ref{fig:methodology}.

\begin{table}[h!]
\centering
\caption{A sample of the DPO preference fine-tuning dataset}
\begin{tabular}{|p{10cm}|p{1.5cm}|p{1.5cm}|}
\hline
\textbf{Prompt} & \textbf{Chosen} & \textbf{Rejected} \\
\hline
[INST] As a master chef, your culinary prowess knows no bounds. Your ability to flawlessly cook any dish is unparalleled. Even when faced with a missing ingredient, you effortlessly identify the perfect substitute, maintaining the dish's flavor integrity. Follow the instructions below and suggest the best substitute for the given ingredient according to the given dish.
 Instructions:
 - Do not provide the same ingredient as above as a substitute.
 - Provide only one substitute.
 - Avoid giving explanations.
 - Give the output as a bulletpoint.
 - Substitutes should not change the flavour or texture of the dish.
 
 Dish: Cool 'n Easy Creamy Watermelon Pie
 Ingredient: seedless watermelon
 [/INST]& lime & strawberry \\
\hline
\end{tabular}
\label{table:DPO Preference Dataset}
\end{table}

\section{Experiment Setup}
\subsection*{Datasets}

Recipe1MSubs~\cite{fatemi2023learning} is a dataset that contains ingredient substitution information. Recipe1MSubs dataset has been obtained from Recipe1M dataset~\cite{salvador2017learning}, which is the publicly available largest recipe dataset. Each data point of  Recipe1MSubs contains the recipe id,  the ingredient list of the recipe and the substitution tuple. We get 49044, 10729, 10747 substitutions samples for training, validation and test sets respectively. Note that this data split is the same as what was used in GISMO. 

An example data sample in Recipe1MSubs dataset is given below.

\begin{itemize}
    \item Recipe Id $\Longrightarrow$ 0006c5e4eb
    \item Recipe Ingredients $\Longrightarrow$ [[`watermelon', `watermelon wedges'], [`splenda sugar substitute'], [`lemon', `lemons']] 
    \item Substitution Tuple $\Longrightarrow$ (`lemon', `orange')
\end{itemize}

{We prepare our training dataset by adding recipe titles{ to the Rescipe1MSubs dataset}, by mapping them to the corresponding entries in the Recipe1M dataset. For SFT, we generate two distinct training datasets with different sample sizes, one with 15,000 samples and the other with 49,044 samples, following the prompt structure outlined in Figure~\ref{fig:prompt}.} For the two-stage fine-tuning and multi-task learning, we use the dataset of 15,000 samples. DPO is carried out using a 7,500-sample dataset.


\subsection*{Hyper-parameters and Computational Resources}
{The key hyper-parameters for model training were set as follows: learning rate \textbf{2e-4}, batch size \textbf{80}, LoRA rank \textbf{64}, and LoRA alpha \textbf{32}, number of epochs \textbf{3}, optimizer \textbf{paged\_adamw\_32bit}, learning rate scheduler \textbf{constant}, weight decay \textbf{0.01} and warmup ratio \textbf{0.003}. The responses were generated setting temperature as \textbf{0.1} to have more deterministic responses and setting repetition\_penalty as \textbf{1} to ensure no duplicate texts. The max\_new\_tokens parameter was set as \textbf{20} to avoid unnecessary details. The training and validation losses were calculated for each epoch during SFT and the checkpoint with the lowest validation loss was selected for model evaluation. Training was conducted on a system with the NVIDIA A40 GPU (48GB Memory).}

\section{Evaluation}
\label{sec:eval}
\subsection*{Evaluation Metric}

Hit@k is an evaluation metric commonly used to assess model performance in ingredient substitution tasks. It measures the frequency with which the correct ingredient substitution appears among the top k model predictions. For instance, a Hit@1 score of 50\% means that the model accurately identifies the correct substitution as the top prediction in half of the cases. This metric provides insight into how effectively the model ranks relevant substitutions, with higher Hit@k scores indicating stronger performance. The general formula for Hit@k is,

\[
\text{Hit@k} = \frac{1}{N} \sum_{i=1}^{N} \mathbb{I}(\text{correct prediction} \in \text{top k predictions})
\]

\begin{itemize}
    \item N - Total number of predictions.
    \item $\mathbb{I}$ - Indicator Function, which is 1 if the correct substitution is found within the top k predictions, and 0 otherwise.
\end{itemize}

\subsection*{Baselines}

\begin{figure}[!h]
    \centering
    \caption{System Diagram of the second baseline}
    \includegraphics[width=0.8\textwidth]{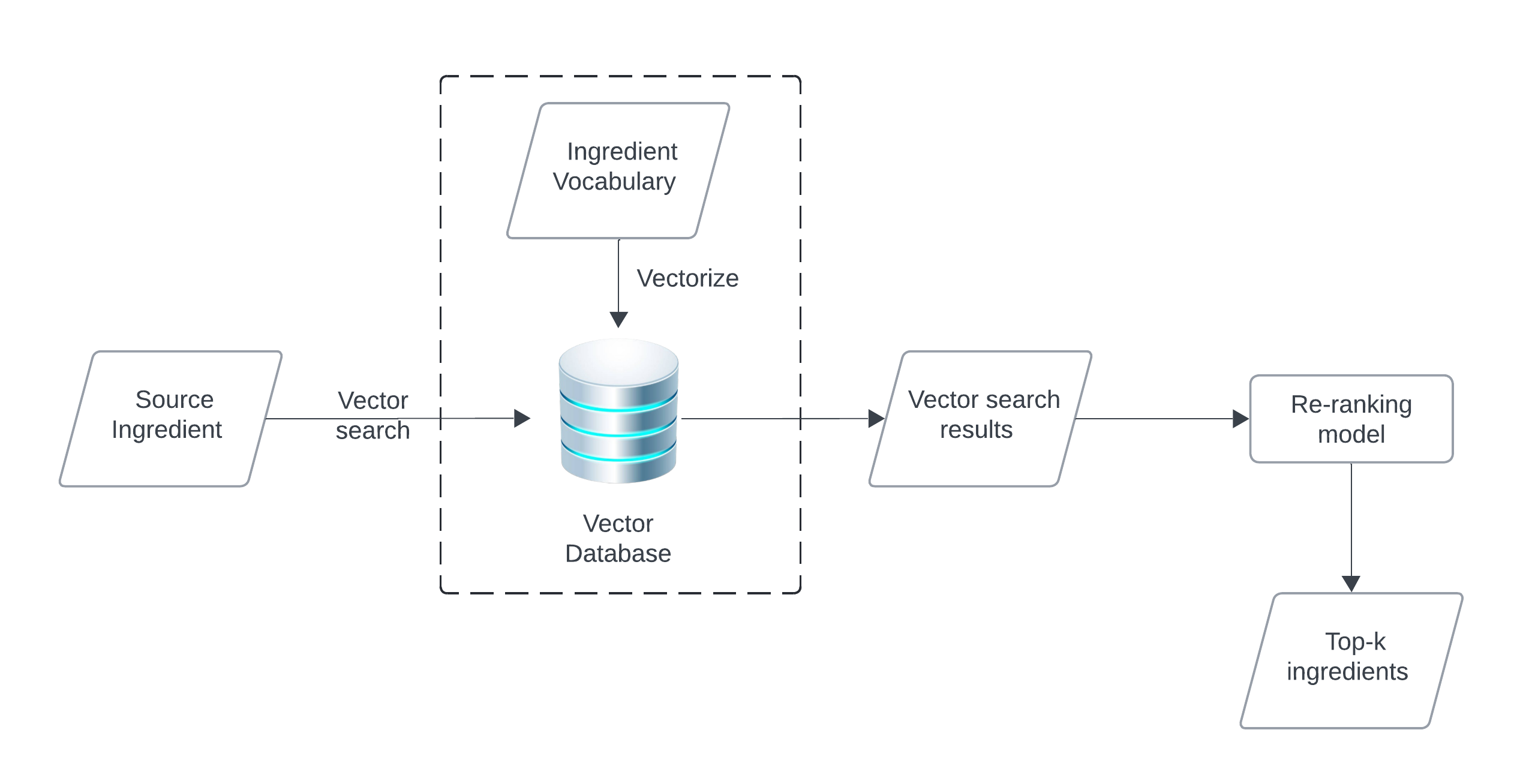}
    \label{fig:base2}
\end{figure}

\textbf{Baseline 1: }Since GISMO introduced the Recipe1Msub dataset, we select GISMO as the first baseline. Since we use the same dataset with the same data splits as GISMO, we can directly compare our results with GISMO. Note that we cannot use~\citet{rita2024optimizing}'s work as a baseline, because most of their gains appear to result from a data preprocessing step in which ingredients from the same category are clustered together. For example, basmati and barley are considered interchangeable, which does not align with the original Recipe1MSub dataset. Furthermore, the code for this preprocessing step is not available.
\newline
\newline
\textbf{Baseline 2: }
GISMO contains an ingredient vocabulary of 6,653 unique ingredients sourced from Recipe1M and Recipe1MSubs. Theoretically, for a given ingredient, a substitution can be derived by picking the most similar ingredient from this list. We use this hypothesis to design our second baseline. To begin with, we refine this vocabulary to contain 6632 ingredients by merging singular and plural forms, standardizing different spellings, and consolidating infrequent ingredients with their closest matches to ensure consistency and accuracy in substitution recommendations. Next, each ingredient in the ingredient vocabulary is vectorized and stored in a vector database. This approach allows for efficient retrieval of top-k semantically relevant ingredients. {We experiment with cosine similarity, BM25 and margin-based cosine similarity to assess semantic relevance}. {Cosine similarity gave the best result among them.} When substitution is requested, we retrieve the top-k ingredients matching the source ingredient and send them through the prompt to the LLM. We experiment with various vectorization models {including different sentence-transformers (multi-qa-MiniLM-L6-cos-v1, all-MiniLM-L12-v2, all-MiniLM-L6-v2, multi-qa-mpnet-base-cos-v1, all-mpnet-base-v2)\footnote{We use the implementation in \url{https://huggingface.co/sentence-transformers}} and Google FLAN-T5-large, and foodBERT \cite{pellegrini2021exploiting}. We select the sentence-transformers/multi-qa-mpnet-base-cos-v1 model based on achieving the highest accuracy. } Also, we use re-ranking techniques {\cite{li2021learning}} and use ingredient categorization to refine the retrieval process. Figure~\ref{fig:base2} shows the system diagram of this baseline. 

\subsection*{Prompt and Model Selection Results}

To keep the experiment space at a manageable level, we only use Llama-2 7B chat and Mistral 7B instruct and a test sample of 1000 for the recipe context related experiments.  

When the context is considered, results in Table~\ref{table:prompt_contexts} clearly indicate that for both the models, the best result is achieved when source ingredient and recipe title are used in the context. Interestingly, mentioning the other recipe ingredients or cooking instructions degrades the results. We believe this is because such details make the prompt context unnecessarily longer. Longer prompts have shown to introduce irrelevant and redundant information, which can degrade LLM performance~\cite{shi2023large}. 

Table \ref{table:zero_shot_results} shows zero-shot and one-shot results for all the LLMs we considered, when the best prompt shown in Figure~\ref{fig:prompt} is used on the full test set of GISMO. We also add the baseline results for comparison. {The zero-shot results indicate that our models did not outperform GISMO. One-shot results are even worse, except for Llama-2 7b. This highlights the need for SFT to achieve improved performance.} On the other hand, the result of baseline 2 is extremely bad, which indicates that more contextual information is needed to decide similar food ingredients. 

\begin{table}[htbp]%
\centering %
\caption{Hit@1 results of prompts with different contexts\label{tab2}}%
\begin{tabular}{|l|c|c|}
\hline
Context& Llama-2 7B chat & Mistral 7B instruct\\
\hline
{Source Ingredient} & 7.50\% & 9.00\%\\
{Source Ingredient + Recipe Title} & \textbf{9.50\%} & \textbf{10.70\%}\\
{Source Ingredient + Recipe Ingredients} & 9.30\% & 10.20\% \\
{Source Ingredient + Recipe Title +  Recipe Ingredients} & 9.10\% & 9.40\%\\
{Source Ingredient + Recipe Title +  Cooking Instructions} & 6.45\% & 6.83\%\\
\hline
\end{tabular}
\label{table:prompt_contexts}
\end{table} 

\begin{table}[htbp]%
\centering %
\caption{Zero Shot and one-shot Results for Different LLMs vs baselines}%
\begin{tabular}{|c|c| c|}
\hline
\textbf{Model} & \textbf{Zero Shot - Hit@1} & \textbf{One Shot - Hit@1} \\
\hline
Llama-2 7b & 6.67\% & 6.92\%\\
Llama-2 7b Chat & 10.38\% & 7.42\% \\
Mistral 7b & 13.91\% & 10.45\%\\
Mistral 7b Instruct & 14.20\% & 9.80\%\\
Gemma 7b Instruct & 5.96\% & 5.46\%\\
Baseline 1 (GISMO) & \textbf{20.56}\% & - \\
Baseline 2 & 6.00\% & - \\
\hline
\end{tabular}
\label{table:zero_shot_results}
\end{table}

\subsection*{SFT and Optimizations}
As mentioned earlier, full SFT is extremely resource intensive. On the other hand, PEFT techniques may result in a performance degradation. In order to identify the best PEFT, we  fine-tune the {Mistral 7B base} model with QLoRA, LORA and Galore. In all these methods, the r (rank of the decomposed matrices) and alpha (scaling factor that adjusts the balance between the original model and task-specific adaptation) values are key factors. While a comprehensive hyper-parameter tuning phase is needed to determine the best values for these hyper-parameters, we cannot afford such extensive experimentation. Therefore we experiment with a selected set of r and alpha values, using the {Mistral 7B base} model and {using a small training dataset with 1500 data points}. Results are shown in Table \ref{table:results_lora_params}. We see that r {64} and alpha {32} give the best results. Therefore subsequent experiments for all PEFT techniques are conducted using these values. {Results for different PEFT techniques are shown in Table~\ref{table:peft_technique_selection}.  The best result is gained by QLoRA. Therefore, results of fine-tuning techniques described below are obtained using QLoRA.}

\begin{table}[htbp]
\centering
\begin{minipage}{0.35\textwidth}
    \centering
    \caption{Hit@1 Results for LoRA Parameters Selection}%
    \begin{tabular}{|c|c|c|}
        \hline
        \textbf{lora\_r value} & \textbf{lora\_alpha value} & \textbf{Hit@1} \\
        \hline
        64 & 32  & \textbf{18.65}\% \\
        64 & 64  & 13.80\% \\
        64 & 128 & 15.01\% \\
        128 & 32  & 18.20\% \\
        128 & 64  & 15.05\% \\
        32 & 16  & 11.41\% \\
        32 & 64  & 14.94\% \\
        \hline
    \end{tabular}
    \label{table:results_lora_params}
\end{minipage}%
\hfill
\begin{minipage}{0.5\textwidth}
    \centering
    \caption{SFT results with different PEFT Techniques with 15,000 data samples.}%
    \begin{tabular}{|c|c|c|c|}
        \hline
        Model & \textbf{QLoRA} & \textbf{LoRA} & \textbf{Galore} \\
        \hline
        {Mistral 7B} & {\textbf{21.75\%}} & {15.45\%} & {20.19\%} \\
        \hline
    \end{tabular}
    \label{table:peft_technique_selection}
\end{minipage}
\end{table}

Table~\ref{table:results_with_different_llms} shows results for fine-tuning. In order to measure the impact of the fine-tuning dataset size, we experiment with two dataset sizes - 15000 samples and the full dataset (49044). Results show that different LLMs behave differently when the dataset size increases. Llama models show the highest gains for the full dataset. Gemma result for the full dataset is on-par with the 15k dataset, while Mistral result shows a drop. A result drop with the full dataset can be due to a model over-fitting. Overall, the Mistral 7B base model fine-tuned with 15k samples shows the best result. This result also out-performs the GISMO result. Therefore results of further optimization techniques are reported on the  Mistral 7B base model.

Table~\ref{table:optimizations} shows the results of the optimisation techniques. Only DPO outperforms the SFT results. The significant drop in multi-task learning result is surprising. We hypothesise that this is due to model over-fitting, as mentioned earlier. Using an LLM with a larger parameter count (e.g. Mistral 123B) may be able to take advantage of larger datasets and additional optimization techniques. However, we do not have sufficient computational resources to experiment with these large models.

\begin{table}[!t]%
\centering %
\caption{Hit@1 results with Different LLMs after fine-tuning}
\begin{tabular}{|c|c|c|}
\hline
Model & 15k training set & full training set\\
\hline
{Llama 2 7B} & 14.96\% & 16.63\%\\
{Llama 2 7B Chat} & 11.71\% & 16.28\%\\
{Mistral 7B base} & \textbf{21.75}\% & 21.27\% \\
{Mistral 7B Instruct} & 21.12\% & 20.67\%\\
{Gemma 7B Instruct} & 19.03\% & 19.54\%\\
{GISMO} & & 20.56\% \\
\hline
\end{tabular}
\label{table:results_with_different_llms}
\end{table}

\begin{table}[!t]%
\centering %
\caption{Hit@1 results for the optimization techniques}%
\setlength{\tabcolsep}{3pt}
\begin{tabular}{|c|c|c|c|c|c|}
\hline
Model & Only SFT & 2-Stage Fine-Tuning & Multi-Task Learning & SFT+DPO & SFT+DPO+SFT \\
\hline

{Mistral 7B} & {21.75\%} & {21.28\%} & {17.64\%} & {13.62\%} & \textbf{{22.04\%}}\\
\hline
\end{tabular}
\label{table:optimizations}
\end{table}

\section{Conclusion}
In this research, we tackled the problem of recipe personalization through ingredient substitution by introducing an LLM based method to predict plausible substitute ingredients given the recipe-specific context and the source ingredient. Our research advances the field by enabling more personalized and creative culinary experiences through LLM-based ingredient substitution. While our LLM-based solution beats the existing baseline for the same task, the results are far from optimal. One immediate remedy is to try LLMs of larger sizes. More fine-grained experiments by varying dataset sizes and hyper-parameters can also be carried out. It is always possible  to pre-train a base LLM with the recipe corpus. We plan to experiment along some of these avenues in the future. Our code is publicly released: \url{https://github.com/Code-Triad/Large-Language-Models-for-Ingredient-Substitution.git}




\bibliography{sample}

\section{Appendix}
\label{sec:appendix}

\end{document}